\documentclass[lettersize,journal]{cas-dc}
\usepackage{amsmath,amsfonts}
\usepackage[authoryear]{natbib}
\usepackage{algorithmic}
\usepackage{adjustbox}
\usepackage{listings}
\usepackage{algorithm}
\usepackage{array}
\usepackage{subcaption}
\usepackage{textcomp}
\usepackage{stfloats}
\usepackage{url}
\usepackage{verbatim}
\usepackage{fvextra}
\usepackage{graphicx}
\usepackage{enumitem}
\usepackage{soul}
\usepackage{xcolor}
\usepackage{ragged2e}
\usepackage{gensymb}
\usepackage{cite}
\usepackage{capt-of}
\usepackage{xcolor}
\usepackage{tcolorbox}
\usepackage{tikz}
\usetikzlibrary{arrows.meta, positioning, shadows, shapes.geometric, shapes.multipart}

\tikzstyle{startstop} = [rectangle, rounded corners, minimum width=3cm, minimum height=1.5cm,text centered, draw=black, fill=red!30]

\tikzstyle{process} = [rectangle, minimum width=3.5cm, minimum height=1cm, text centered, draw=black, fill=blue!30]
\tikzstyle{arrow} = [thick,->,>=stealth]
\newcommand{\red}[1]{\textcolor{black}{#1}}

\lstdefinelanguage{json}{
    basicstyle=\ttfamily\footnotesize,
    breaklines=true,
    numbers=none,
    stepnumber=1,
    showstringspaces=false,
    tabsize=2
}

\lstdefinestyle{textlist}{
    basicstyle=\ttfamily\scriptsize,
    breaklines=true,
    numbers=none,
    showstringspaces=false
}
\newcounter{numquote}
\newenvironment{lquote}
    {\refstepcounter{numquote} 
    \par\noindent Response \arabic{numquote}:}  
    {\par} 

\begin{document}
\let\WriteBookmarks\relax
\def\floatpagepagefraction{1}
\def\textpagefraction{.001}

\title[mode = title]{Instructor-Worker Large Language Model System for Policy Recommendation: a Case Study on Air Quality Analysis of the January 2025 Los Angeles Wildfires}
\shortauthors{K. Gao et al.}  
\author[1]{Kyle Gao}[orcid=0000-0002-8320-6308]
            \credit{Conceptualization, Methodology, Validation, Formal analysis, Investigation, Data Curation, Writing - Original Draft, Visualization}
\ead{y56gao@uwaterloo.ca}
\affiliation[1]{organization={Department of Systems Design Engineering},
            addressline={University of Waterloo}, 
            city={Waterloo},
            postcode={Ontario, N2L 3G1}, 
            country={Canada}}
            
\author[1]{Dening Lu}[orcid=0000-0003-0316-0299]
\credit{Investigation, Writing - Review \& Editing}
\ead{d62lu@uwaterloo.ca}            

\author[3]{Liangzhi Li}
\credit{Investigation, Writing - Review \& Editing}
\ead{liliangzhi@chd.edu.cn}
\affiliation[3]{organization={College of Land Engineering},
            addressline={Chang'an University},
            city={Xi'an},
            postcode={Shanxi, 710064},
            country={China}}
\author[4]{Nan Chen}
\credit{Investigation, Writing - Review \& Editing}
\ead{chcdut@126.com}
\affiliation[4]{organization={School of Computer Science},
            addressline={Xi'an Aeronautical University},
            city={Xi'an},
            postcode={Shanxi, 710087},
            country={China}}
            
\author[2]{Hongjie He}
\credit{Investigation, Writing - Review \& Editing}
\affiliation[2]{organization={Department of Geography and Environmental Management},
            addressline={University of Waterloo}, 
            city={Waterloo},
            postcode={Ontario, N2L 3G1}, 
            country={Canada}}
\ead{hongjie.he@uwaterloo.ca}
\cormark[1]
            
\author[1]{Jing Du}
\credit{Investigation, Writing - Review \& Editing}
\ead{j7du@uwaterloo.ca}       

\author[1,5]{Linlin Xu}[orcid=0000-0002-3488-5199]
            \credit{Funding Acquisition, Resources, Writing - Review \& Editing, Supervision}
\ead{lincoln.xu@ucalgary.ca}
\cormark[2]

\affiliation[5]{organization={Department of Geomatics Engineering},
            addressline={University of Calgary},
            city={Calgary},
            postcode={Alberta, T2N 1N4},
            country={Canada}}

\author[1,2]{Jonathan Li}[orcid=0000-0001-7899-0049]
\credit{Funding Acquisition, Resources, Writing - Review \& Editing, Supervision}
\ead{junli@uwaterloo.ca}

\cortext[1]{Corresponding author hongjie.he@uwaterloo.ca, (Hongjie He)}
\cortext[2]{Corresponding author lincoln.xu@ucalgary.ca, (Linlin Xu)}

\begin{keywords}
ChatGPT, Large Language Models, GIS, air quality, wildfire, natural disaster, \red{ChatGPT}
\end{keywords}
\maketitle
\begin{abstract}
The Los Angeles wildfires of January 2025 caused more than 250 billion dollars in damage and lasted for nearly an entire month before containment. Following our previous work, the Digital Twin Building, we modify and leverage the multi-agent Large Language Model (LLM) framework as well as the cloud-mapping integration to study the air quality during the Los Angeles wildfires. 
Recent advances in large language models have allowed for out-of-the-box automated large-scale data analysis. We use a multi-agent large language system comprised of an Instructor agent and Worker agents. Upon receiving the users' instructions, the Instructor agent retrieves the data from the cloud platform and produces instruction prompts to the Worker agents. The Worker agents then analyze the data and provide summaries. The summaries are finally input back into the Instructor agent, which then provides the final data analysis. We test this system's capability for data-based policy recommendation by assessing our Large Language Model System with Instructor-Worker Architecture's health recommendations and numerical summarizations based on the air quality data during the Los Angeles wildfires. 
\end{abstract}

\section{Introduction}
Recent Large Language Models (LLMs) have shown tremendous progress and achieved human expert-level performance across a multitude of tasks in science and engineering. In the era of Internet-of-Things (IoT) connected data streams and intelligent agent-like large language models, data processing, data analytics, and data-based problem solving become ever more accessible. 

We propose an LLM System with Instructor-Worker Architecture for geographic information system data analysis. In this system, the Instructor LLM is responsible for interpreting user prompts. It generates Application Programming Interface (API) calls to the cloud platform to retrieve data. It also generates instructions for the Worker LLMs whose roles are to analyze and summarize the data. The Workers then provide the Instructor with the data summary. The Instructor finally analyzes the data according to the initial user prompt. 

This system is based on our earlier work, the Digital Twin Building framework \citep{gao3}, which integrates 3D meshes of buildings \citep{gbm,gao_3dgs1} with cloud-connected database and data streams from Google Maps Platforms with a Multi-Agent LLM System. Our contributions are as follows:
\begin{itemize}
    \item Improving on our previous multi-agent LLM system, we create a \textbf{new LLM System with Instructor-Worker Architecture} with integrated prompt and response passing between LLMs, and automated cloud-based data retrieval using LLM generated query code.
    \item We show the system's capability for Geographic Information Systems (GIS) data analysis for policy recommendation scenarios such as health recommendations and report generation.
    \item We perform a case study based on the air quality data for the January 2025 Los Angeles Wildfires in the Pacific Palisade area.
\end{itemize}
Our study aims to integrate a large language model framework with a cloud-based GIS API to enable automated analysis of real-time air quality and other \red{GIS} data. Rather than focusing on raw satellite data processing, our system leverages the Google Maps Platform, which aggregates sources like ground-based sensors and remote sensing platforms such as MODIS and SEVIRI, allowing LLMs to retrieve and interpret live environmental metrics through API access.

\section{Background and Related Work}

\subsection{Large Language Models}
Large Language Models (LLMs) are Transformer-based neural networks \citep{transformer} trained on vast text and image datasets, primarily sourced from web crawls. Designed for Natural Language Processing (NLP), they process text inputs and generate human-like responses. Some models, such as "DeepseekV3/R1" \citep{deepseekv3, deepseekr1}, support Optical Character Recognition (OCR), while others, like "ChatGPT-4o", "Gemini" models, "Llamma3.2+" \citep{gpt4, gemini, llama3}, can fully analyze image content. Since 2023, LLMs have gained widespread adoption, demonstrating expert-level problem solving in scientific and engineering fields \citep{gpqa, math500}. Due to their large size, local deployment is often impractical. Instead, providers like OpenAI and Deepseek offer web-based interfaces as well as APIs, allowing large-scale data processing with minimal hardware, even on mobile devices.

\subsection{Large Language Models in GIS}
Frameworks have been developed to give LLM geographic information science and systems (GIS) data analysis capabilities or to build LLM-based geographic information systems with code generation-execution capabilities. ToolLLM \citep{2023toolllm} is a lightweight, agile framework that augments language models with external tool integration. LangChain \citep{2023langchain}, on the other hand, offers a modular, composable architecture, enabling developers to seamlessly chain together LLM prompts, data ingestion pipelines, and auxiliary utility functions. These frameworks are external libraries that can be used to build LLM systems. Specific to GIS, LLM-Geo \citep{llmgeo} is an early example of autonomous GIS with code generation capabilities for data analysis, but lacking data retrieval capabilities. \citet{2024llmgis} designed a system to retrieve API calls using Retrieval Augmented Generation for \red{GIS} data analysis with local LLM models, and benchmarked for function call performance.

\subsection{Google Maps Platform}
Google Maps Platform is a cloud-based service within Google Cloud that offers APIs for GIS, mapping, and remote sensing. As a cloud-based mapping service, it provides elevation data, geocoding, building polygon retrieval, map retrieval, and route planning. It also provides APIs for retrieving air quality data, solar data, and plant pollen data.  Though not as widely recognized as Google Earth Engine in the GIS community, it is used for a range of applications like navigation, object tracking, city modeling, and \red{GIS data} analysis \citep{maps1,maps2,maps3,maps4, gao3}. Additionally, the platform provides APIs for real-time traffic, solar potential, air quality, and pollen data, and is integrated into various commercial cloud-based mapping solutions.

\subsection{GIS-based and Deep Learning-based Air Quality Analysis}
Many commercial/institutional GIS providers and platforms have air quality data integration sourced from a variety of ground-based and remote-sensing-based sensors. There are multiple studies on the statistical analysis of air quality data, as well as predictive modeling, and spatial distribution modeling using classical methods (non-AI based) methods \citep{air1,air2,air3,air4}.

Deep-learning-based air quality forecasting has recently been popularized \citep{dlair_review}. A combination of architectures leveraging convolutional networks and modules \citep{cnnair,cnnair2} for spatial data featurization and learning, along with recurrent neural network (RNN) family models for temporal information learning, is commonly used. For pure time series forecasting, RNN family models \citep{rnnair1,rnnair2,rnnair3} or newer Transformer models \citep{transformerair1,transformerair2} are often employed. These are typically supervised time series or geospatial-temporal forecasting models with no GIS integration.

Even more recently, LLM-based air quality analysis has emerged. \citet{llmair2} proposed a concept to introduce GPT2-base edge computing servers for local air quality sensors for PM2.5 prediction. \citet{llmair} fine-tuned an LLM for PM2.5 prediction. However, these LLM-based prediction models have subpar performance since LLMs are in general more adapted to semantic analysis than mathematical modeling. This is corroborated by their benchmarks which show that the errors are on the same order of magnitude as the predicted PM2.5 levels. 

\citet{llmchatbotair} proposed an LLM-chatbot framework with multiple local LLM models for air quality analysis and code generation, capable of querying an in-house air quality dataset with over 7 years of data and producing insights. This was tested against a manually compiled set of questions with user-written code. However, the authors did not detail their code execution.

The works of \citet{llmchatbotair} and \citet{llmgeo} are most relevant to ours. Like \citet{llmgeo} we use cloud-hosted LLMs, enabling lightweight mobile deployment. However, unlike \citet{llmgeo}, our system is agentic and uses multiple LLMs which pass system prompts and messages to each other. Our framework assigns simple tasks to lightweight agents and reserves complex tasks for more capable ones. We include a Code Execution Module with compliance checks to constrain outputs to pre-approved functions. Unlike \citet{llmchatbotair}, our system retrieves data from cloud sources such as Google Maps, whereas theirs does not. We also differ in hosting: their models run locally, and ours offload computation to remote servers.

\citet{llmchatbotair} and \citet{llmgeo} focus on Python code generation, while we focus on data analysis. Their LLM executes and debugs code freely, while ours enforces strict compliance through system prompts. We also avoid orchestration libraries like LangChain \citep{2023langchain} and ToolLLM \citep{2023toolllm}, keeping our framework lightweight and portable.


\section{Method} \label{sec:methodology}
\subsection{Google Maps Platform Air Quality API}
Google Maps Platform offers air quality data in the form of current air quality, 720-hour history at 500 m resolution, and current air quality tile maps. This data is sourced from a variety of local air quality monitoring stations and satellite sensors from local providers and interpolated when necessary. Google Maps Platform offers air quality measurements for six pollutants: carbon monoxide, nitrogen dioxide, ozone, particulate matter $< 10$ $\mu m$, fine particulate matter $<$ 2.5 $\mu m$, and sulfur dioxide. The API also provides information on potential pollutant sources, health hazards, local air quality index and health advisory warnings and recommendations. 

We limit the scope of our study to Google Maps Platform as an API-accessible cloud-GIS data source and data stream. However, with minor modifications, other API-accessible cloud-GIS platforms' data sources, such as OpenAQ \citep{openAQ} can be integrated into our framework for data retrieval.

\subsection{Case Study: Los Angeles Wildfires}
The Pacific Palisades Region of Los Angeles, California was affected by severe wildfires in January 2025, starting on January 7th and rapidly spreading over the next few days. It was contained after 24 days on January 31st. \red{It was among the two principal wildfire incidents that ravaged Los Angeles in January 2025. The other major event was the Eaton wildfires. These disasters caused damage exceeding 250 billion dollars, and consumed an expanse of nearly 155 km$^2$.} 

Large wildfires are known to produce inhalable particulate matter at a level significantly above the baseline of a region. Two common measures of these particulates are PM2.5 level and PM10 level, which stand for particulates of 2.5 $\mu m$ or less and particulates of 10 $\mu m$ or less \citep{lafires}

\begin{figure*}[h!]
\centering
        \includegraphics[width=0.95\textwidth]{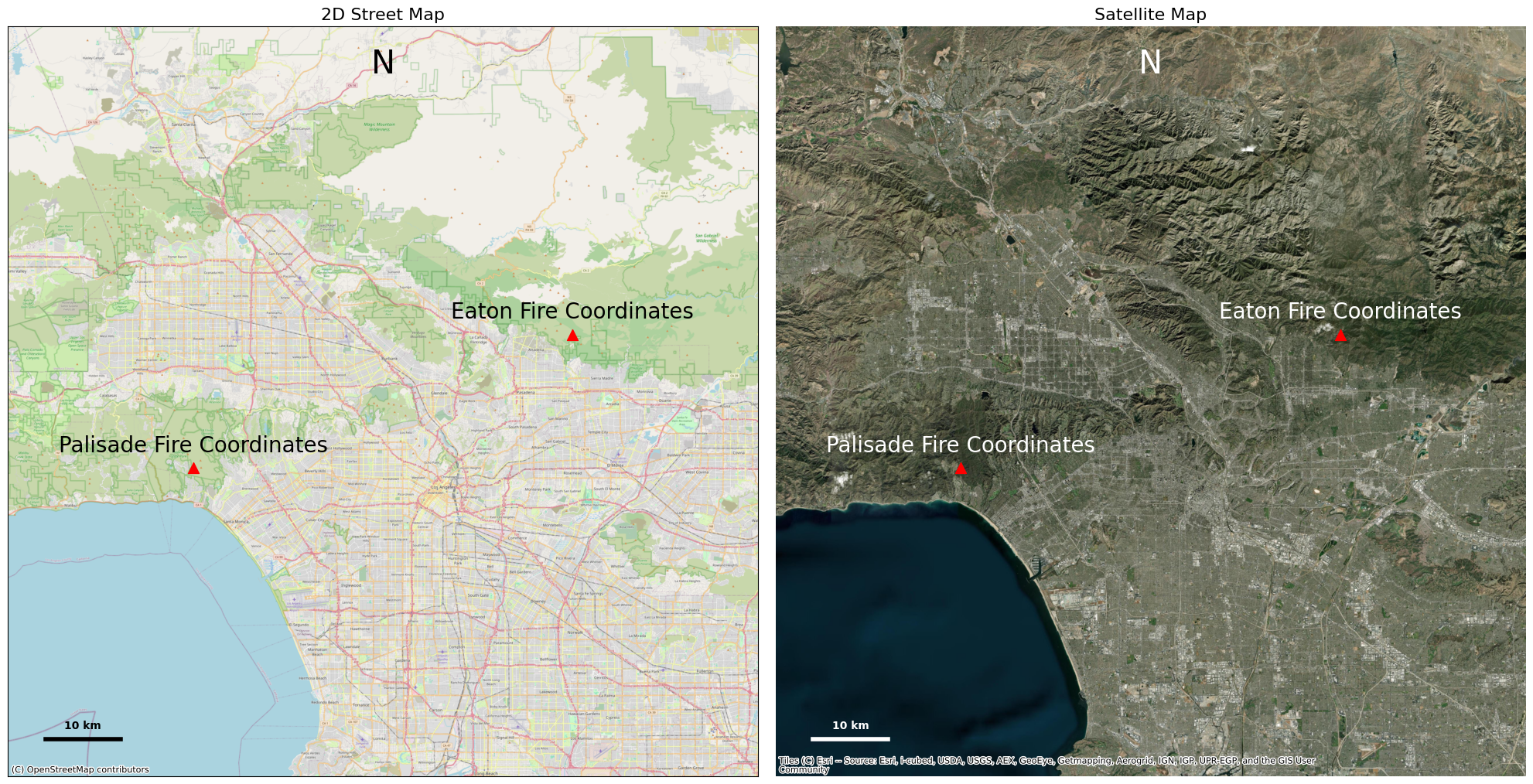} 
\caption{\red{2D street map and satellite map of the region of study with data retrieval locations marked. The satellite map is tiled from multiple sources (image attribution at the bottom right). The 2D street map is retrieved from OpenStreetMap \citep{osm}.}}\label{fig:region_of_study}
\end{figure*}

To analyze the potential of the Multi-Agent LLM system for policy recommendation, we retrieved 30 days of air quality data from Google Cloud Platforms using the LLM system's query. The entire dataset is too large for the Instructor LLM or the Worker LLM to analyze at once. Therefore, it was retrieved from the cloud and processed as chunks. The entire 720-hour dataset was separated into 30 chunks of 24 hours to which we appended the 24-hour data of the previous day, resulting in 30 chunks of 48 hours.\footnote{We added the second day's data to the first chunk. For other chunks, we added the previous day's data.} We then used the LLM system to recommend daily health advisory warnings given the air quality data. This data also contains ground truth health advisory warnings, which we benchmark against. 

\red{We retrieved the air quality data for both the Palisades (Topanga State Park, 34.0725° N, 118.5445° W) and Eaton (Angeles National Forest, 34.2050° N, -118.0880° W) neighborhoods as shown in Figure \ref{fig:region_of_study}. However, this study focuses on the data from the Palisades wildfires.}

\subsection{Multi-agent Analysis of Air Quality} 

\begin{figure}[htpb]
\centering
        \includegraphics[width=0.5\textwidth]{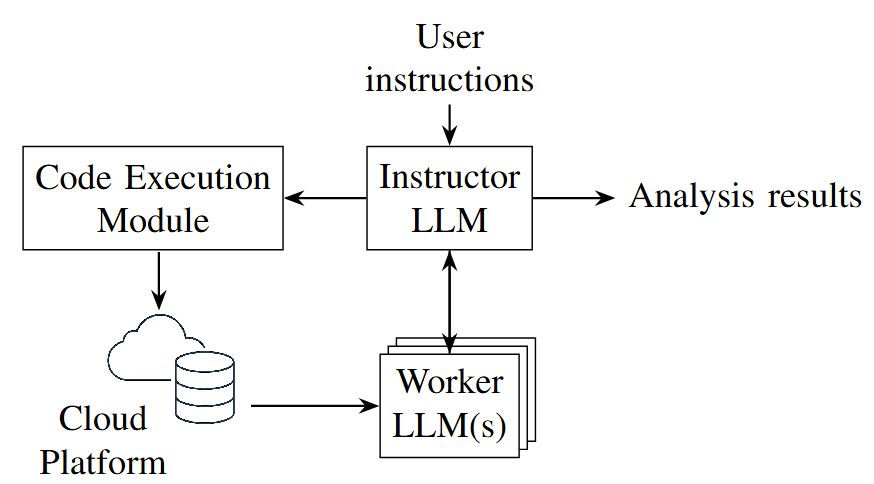} 
\caption{Instructor-Worker System for cloud-based datasets and data streams.}\label{fig:instructor_worker}
\end{figure}

We designed an \textbf{LLM System with Instructor-Worker architecture} adapted from the Multi-Agent LLM system in Digital Twin Buildings \citep{gao3}. The framework of our new system is illustrated in Fig. \ref{fig:instructor_worker}. The Instructor LLM is a Chain-of-Thought reasoning model such as Deepseek-R1/GPT-O1 with a higher problem-solving capability, but a more expensive API call price and typically slower response time. The Worker(s) can either be reasoning models or autoregressive (chat-style) models that are inexpensive to use and have a faster response time than reasoning models.

\noindent
\textbf{Code Execution Module: }The Instructor LLM is responsible for analyzing the initial user input. We created a library of system prompts for useful functions and API calls from our Python wrapper introduced in Digital Twin Building's Google Maps Platform integration \citep{gao3}. From a predetermined list, the Instructor chooses the functions required given the user instructions. For the Code-Execution LLM, we then load the user's instruction as a \textit{user prompt}, the Instructor LLM's instruction as a \textit{system prompt}, and the appropriate system prompts containing documentation for the chosen functions as an additional \textit{system prompts}.

The Code-Execution LLM's response string is processed through two code compliance checks and a string-to-code execution function. For the first check, the framework initializes another LLM agent and asks it to examine the response string for any malicious code. For the second check, string formatting is used to check whether the response string is compliant with the initial documentation. If both checks are passed, the API call from the Instructor LLM is executed as code using Python's built-in string execution function \textit{exec()}.

\textbf{Data Analysis:} Although we only use the Code Execution Module for the Google Maps Platform API call for this case study, it can also be used to post-process the final analysis, e.g., to generate plots during analysis. For air quality data analysis based on data summarization, the Instructor LLM then provides a set of instructions for the Worker LLM. The Worker's goal is to summarize the dataset into as few tokens as possible. This summary is then returned as context input into the Instructor model for dataset analysis.

\subsection{Evaluation Metrics}
The BERTScores \citep{bertscores} (BERT precision, recall, and F1 score) are used to assess text generation results in the presence of a reference text. It is highly correlated with human judgment and captures semantically meaningful similarities. The BERTScores use a pretrained BERT model \citep{bert} to embed both an input text $(c_1,c_2,...,c_m)$ and a ground truth text $(r_1,r_2,...r_n)$, as $\mathbf{e}_{c_i} = \text{BERT}(c_i), \quad \mathbf{e}_{r_j} = \text{BERT}(r_j)$. Then, a cosine similarity matrix is calculated. BERTScore precision, recall, and F1 are then computed based on similarity matching.

The cosine similarity is given by
\begin{equation} \label{eq:cosine}
s_{ij} = \frac{\mathbf{e}_{c_i} \cdot \mathbf{e}_{r_j}}{\|\mathbf{e}_{c_i}\| \|\mathbf{e}_{r_j}\|},
\end{equation}
 where $i$ indexes candidate text (model output) and $j$ indexes reference text (ground truth).
 
The BERTScore precision is obtained by matching each candidate token with the most similar reference token and averaging their mutual cosine similarities. 
\begin{equation} \label{eq:bertp}
P_{BERT} = \frac{1}{m} \sum_{i=1}^{m} \max_{j} s_{ij}.
\end{equation}
The BERTScore recall is obtained by matching each reference token with the most similar candidate token and averaging their mutual cosine similarities. 
\begin{equation} \label{eq:bertr}
R_{BERT} = \frac{1}{n} \sum_{j=1}^{n} \max_{i} s_{ij}.
\end{equation}
The BERTScore F1 is calculated using the harmonic mean of BERTScores precision and recall.
\begin{equation}\label{eq:bertf1}
F_{1BERT} = 2 \cdot \frac{P_{BERT} \cdot R_{BERT}}{P_{BERT} + R_{BERT}}.
\end{equation}   

\section{Experiments and Discussion}
\subsection{Preliminary Analysis} \label{multi_agent_analysis}
\begin{figure}[htbp]
    \centering
    \begin{subfigure}{0.5\textwidth}
        \centering
        \includegraphics[width=\textwidth]{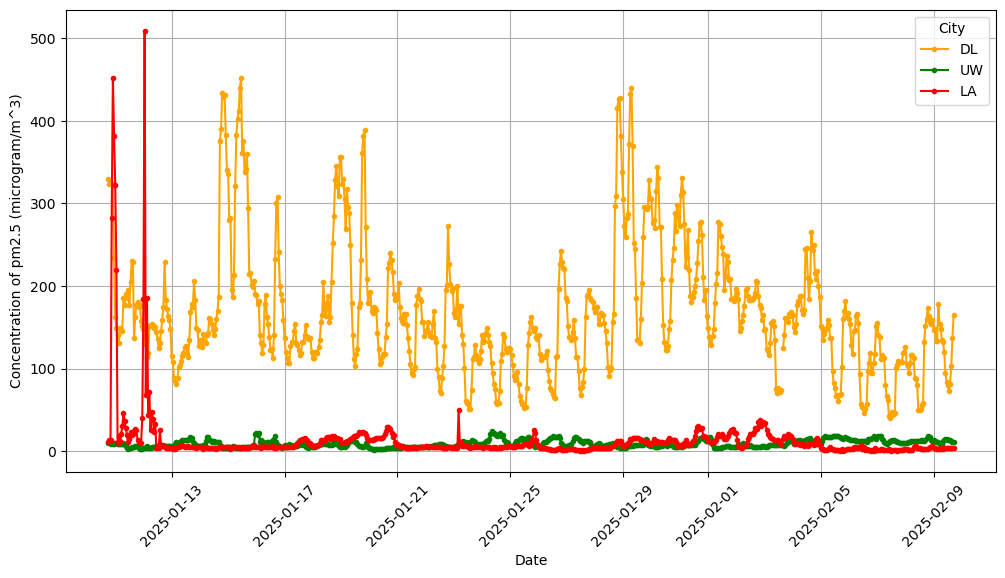}  
    \end{subfigure}
    
    \begin{subfigure}{0.5\textwidth}
        \centering
        \includegraphics[width=\textwidth]{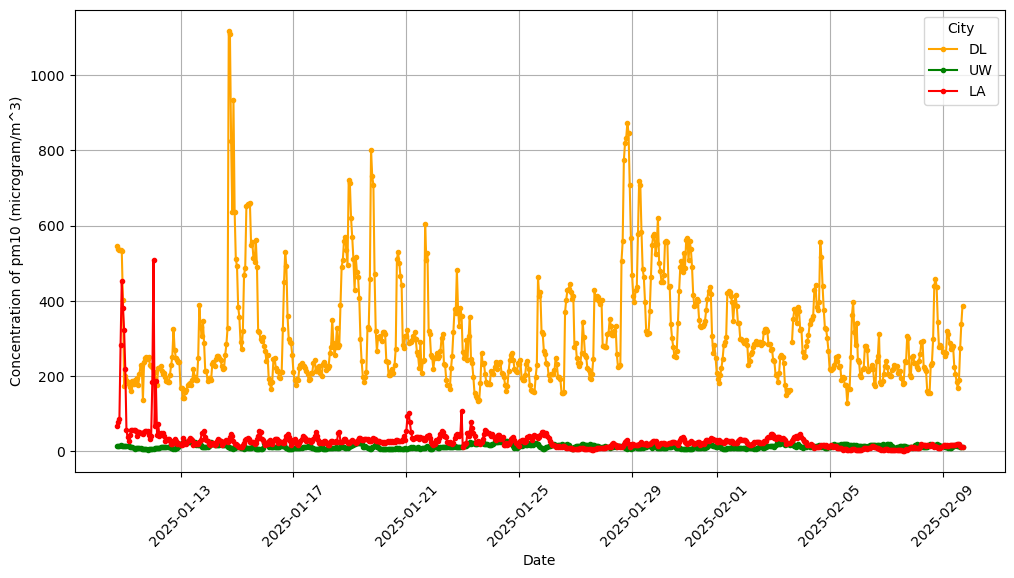}  
    \end{subfigure}
    \caption{PM2.5 (Top) and PM10 (Bottom) levels from January 9th to February 9th in select neighborhoods in Los Angeles (LA),  Waterloo (UW), Delhi (DL).}\label{fig:pm_levels}    
\end{figure}

To analyze the impact of the wildfires on the Pacific Palisades' PM2.5 and PM10 levels, we first use the instructor LLM to retrieve the 720 hour air quality history of the Topanga State Park neighborhood, the suspected starting point of the Palisades wildfire. The data begins on January 9th 2025 due to the 720 hours limit at the time of data retrieval. We also retrieve the air quality at the University of Waterloo in Waterloo, Canada, as a low pollutant baseline, and a Delhi neighborhood in New Delhi, Delhi, India, as a high pollutant baseline. The PM2.5 and PM10 levels of the respective neighborhoods (Table \ref{tab:airqualityneighborhood}) are plotted in Figure \ref{fig:pm_levels}. 
\begin{table}[htpb]
\centering
\scriptsize
\caption{\red{High and low air quality baseline comparison neighborhoods coordinates}}\label{tab:airqualityneighborhood}
\begin{tabular}{c|c|c}
\hline
City & Coordinates & Neighborhood \\ \hline
Los Angeles, USA & 34.0725° N, 118.5445° W & Topanga State Park \\ 
Waterloo, Canada & 34.07022° N, 80.5449° W & University of Waterloo \\ 
New Delhi, India & 28.7041° N, 77.1025° E & Abhinav Public School \\ \hline
\end{tabular}
\end{table}

When examining the Los Angeles Palisade PM2.5 and PM10 levels, two major spikes can be observed, roughly on January 9-10th and 11-12th. There are also a multitude of smaller spikes. The January 9-10th and 11-12th major spikes have amplitudes which are significantly above the Waterloo baseline for clean air but are comparable to the baseline of Delhi. Surprisingly, the LA PM2.5 and PM10 pollutant levels are fairly comparable to Waterloo's levels outside of the spike events, despite the ongoing wildfires. This suggested a certain degree of local-ness to the pollution caused by wildfires.

Figure \ref{fig:pm_levels} shows two major spikes in particulate matter in the first three to four days of the time series, as well as a few minor spikes later on. The particulate matter levels returned to baseline fairly quickly. The baseline air quality near the measurement site is close to clean city baseline levels despite the ongoing fires that lasted for the entire month. The data seem to show that PM2.5 and PM10 levels are highly localized, perhaps with spikes occurring only with nearby active wildfires close to the sensor site. This effect seems to not be mentioned in the literature and should be investigated further.

Below is an example of the framework in use when the user asks for an analysis of the PM2.5 and PM10 levels of the LA neighborhood during the time period. The Instructor LLM provides the instructions to the Worker LLMs. 
\vspace{12pt}
\begin{lquote}
Example Instructions for Worker LLMs.
\hrule
1. \textbf{Data Extraction}: For each day in the dataset, calculate the daily average levels of PM2.5 and PM10 using the hourly data.  

2. \textbf{Outlier Detection}: Identify outlier events by flagging days where PM2.5 or PM10 levels exceed 1.5 times the interquartile range (IQR) above the 75th percentile or below the 25th percentile.  

3. \textbf{Summarization}: Compress the data into a structured format:  
   - Daily averages: \{date: YYYY-MM-DD, PM2.5\_avg: value, PM10\_avg: value\}  
   - Outlier events: \{start\_date: YYYY-MM-DD, end\_date: YYYY-MM-DD, pollutant: PM2.5/PM10, max\_level: value\}  

4. \textbf{Output}: Return only the structured summary without additional commentary or explanations.
\hrule
\end{lquote}
\vspace{12pt}

Then, Worker LLMs respond with their summary for their respective chunk. An example response for a chunk is provided as follows.
\begin{quote}
\tiny
\begin{tcolorbox}[colback=white, colframe=black, boxrule=0.5pt, sharp corners]
\begin{lstlisting}[language=json]
  "daily_averages": [
    {"date": "2025-02-04", "pm25_avg": 8.94, "pm10_avg": 13.23},
    {"date": "2025-02-05", "pm25_avg": 2.48, "pm10_avg": 7.34},
    {"date": "2025-02-06", "pm25_avg": 2.63, "pm10_avg": 7.37},
    {"date": "2025-02-07", "pm25_avg": 1.56, "pm10_avg": 5.63},
    {"date": "2025-02-08", "pm25_avg": 3.12, "pm10_avg": 13.12},
    {"date": "2025-02-09", "pm25_avg": 3.34, "pm10_avg": 15.12}
  ],
  "outlier_events": [
    {"start_date": "2025-02-04", "end_date": "2025-02-04",
     "peak_pm25": 16.2, "peak_pm10": 18.44},
    {"start_date": "2025-02-08", "end_date": "2025-02-09",
     "peak_pm25": 5.99, "peak_pm10": 19.79}
  ]
\end{lstlisting}
\end{tcolorbox}
\end{quote}
The summaries for all chunks are aggregated and passed back to the Instructor LLM, which analyzes them according to user instructions. For example, when asked to analyze the time period for outliers and possible causes, the Instructor LLM responds with the following.
\vspace{12pt}
\begin{lquote}
Key Pollutant Spike Events \& High-Pollution Days Analysis
\hrule
\textbf{1. Extreme Pollution Events (Significantly Above Regional Averages)}  

- \textbf{January 10, 2025}:  
  - \textbf{PM2.5}: Peaked at \textbf{382.03} (extreme spike).  
  - \textbf{PM10}: Peaked at \textbf{452.24} (extreme spike).  
  - \textbf{Likely Event}: Severe short-term pollution event (e.g., industrial accident, wildfire, or dust storm).  

- \textbf{January 12, 2025}:  
  - \textbf{PM2.5}: Peaked at \textbf{509.07} (highest recorded spike).  
  - \textbf{PM10}: Peaked at \textbf{509.07} (unusually high co-spike).  
  - \textbf{Likely Event}: Prolonged combustion event (e.g., large-scale biomass burning or industrial emissions).  

- \textbf{January 21–22, 2025}:  
  - \textbf{PM10}: Spiked to \textbf{92.45} (Jan 21) and \textbf{108.05} (Jan 22).  
  - \textbf{Likely Event}: Localized dust storms or construction activity.  

- \textbf{February 2, 2025}:  
  - \textbf{PM2.5}: Peaked at \textbf{37.86}; \textbf{PM10} at \textbf{45.10}.  
  - \textbf{Likely Event}: Moderate industrial/vehicular pollution surge.  

\textbf{2. Days with Sustained High Pollution}  

- \textbf{January 10–16, 2025}:  
  - \textbf{PM2.5}: Consistently elevated (avg \textbf{166.91} on Jan 10, dropping to \textbf{4.92} by Jan 13).  
  - \textbf{PM10}: Similarly high (avg \textbf{174.79} on Jan 10, stabilizing at \textbf{$\sim$ 27.92} post.  
  \hrule
\end{lquote}
\vspace{12pt}
This LLM system was correctly able to identify outlier time periods and provide a probable cause for the level of pollutants, as well as assess that the pollutant levels were stabilized at around 10\% of the maximum level. We should note that both PM10 and PM2.5 levels peaked at \textbf{509.07 $\mu$g/$m^3$}. This number was not hallucinated by the LLMs and was verified against the raw data from the Google Maps Platform. \red{It is improbable that two measurements exhibit peaks that coincide to five significant digits. This concurrence may possess statistical or instrumental significance, for example, arising from inherent sensor resolution constraints}. The smaller spikes of PM10 and PM2.5 pollutants after stabilization confused the LLM system (e.g., January 21-22). It did not deduce that the region was devastated by large-scale wildfires throughout the entire month: despite the fires moving away from the sensor after the initial few days, the major cause of the pollutants was still the wildfire and not dust storms or construction.

\subsection{Multi-Agent Health Recommendation and Benchmark}
Google Maps Platform Air Quality API provides health recommendations for healthy people, people with lung diseases, heart diseases, and pregnancies, as well as the elderly and children. To compare the effectiveness of the Instructor-Worker framework for policy recommendation, we prompted the Instructor to generate health recommendations for healthy people and people with lung diseases for the following three days in the Topanga State Park neighborhood in Los Angeles: January 10 (low air quality), January 22 (moderate air quality), and February 9 (good air quality). Although these are not official health categorizations from the World Health Organization (WHO) or the Centers for Disease Control (CDC), we prompted the LLMs with these Google Maps Platform health classes to conform with our ground truth. Example responses for three different iterations are shown below. 
\vspace{12pt}
\begin{lquote}\label{verb:recommendations}
January 10 Health Recommendation for People with Lung Disease:  
\hrule
   - \textbf{Iteration 1}: Air quality is unhealthy and may aggravate lung conditions.  
     - Advised to avoid all outdoor physical activities.  
     - Stay indoors in a well-ventilated space.  
     - Consult a healthcare provider if symptoms arise.  

   - \textbf{Iteration 2}: Air quality is hazardous.  
     - Remain indoors and keep activity levels low.  
     - Follow medical advice and consider using air purifiers.
     
...

   - \textbf{Iteration 4}: Air quality is very unhealthy.  
     - Avoid all outdoor activities.  
     - Stay indoors with windows closed.  
     - Use air purifiers if available.  
     - Follow doctor's recommendations and have quick-relief medication on hand. 
     
...

   - \textbf{Ground Truth}:  
     - Reduce the intensity of outdoor activities.  
     - Keep relevant medications available and consult a doctor if needed.  
     - Limit exposure to busy roads, open fires, and smoke sources.  
     - Reduce time near industrial emission stacks.  
     - Staying indoors with an activated air filtration system is recommended for long-term health.  
\hrule
\end{lquote}
\vspace{12pt}

\begin{figure}[htbp]
    \centering
    \begin{subfigure}{0.45\textwidth}
        \centering
        \includegraphics[width=\textwidth]{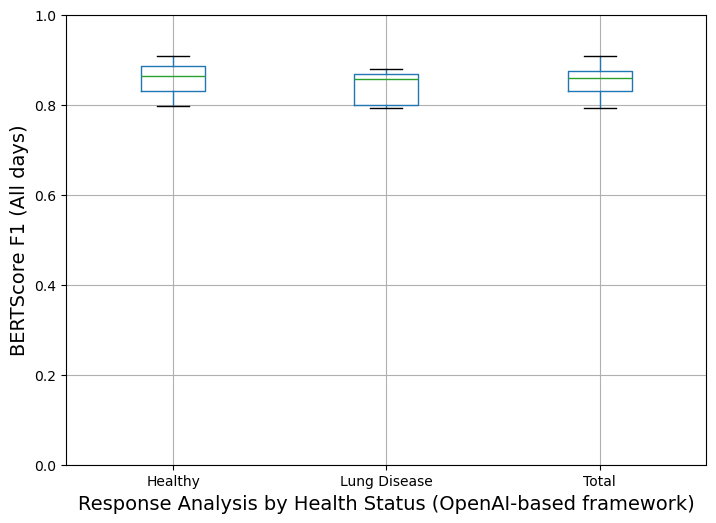}  
        \subcaption{BERTScore F1 for health recommendation comparison for healthy people vs. people with lung disease.}\label{fig:box_plotsberta}    
    \end{subfigure}
    
    \begin{subfigure}{0.45\textwidth}
        \centering
        \includegraphics[width=\textwidth]{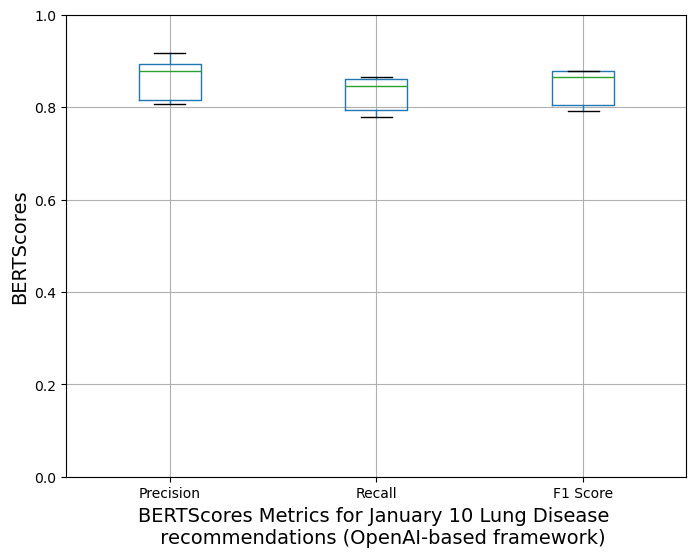}  
        \subcaption{BERTScore F1 score, precision, recall for January 10 recommendation for people with lung disease.}\label{fig:box_plotsbertb}    
    \end{subfigure}
        \caption{BERTScore distributions boxplots. Top: comparison of healthy people, people with lung disease, and the total population. Bottom: BERTScore F1, precision and recall recommendations for people with lung disease on a high pollution day.}\label{fig:fig3}    
\end{figure}

Despite using an LLM temperature of zero, the framework produced semantically similar but structurally different sentences. We repeated 10 iterations of the framework with identical initial instructions to assess the variance in the responses, which were compared against ground truth recommendations from the Air Quality API using the BERTScores (\ref{eq:bertp}-\ref{eq:bertf1}). \textbf{Iteration 4} is shown above as the iteration with the lowest scores.

Figure \ref{fig:box_plotsberta} shows a fairly large within-(sub)class variance, which is comparable to the total variance (for people with lung disease). When plotting the BERTScores for health recommendations for people with lung disease on January 10 in Figure \ref{fig:box_plotsbertb}, we observe a rather large interquartile range of around 0.6 for all three scores. This is comparable to the interquartile range for people with lung disease across all days and is larger than the total interquartile range. This suggests that the variance is primarily driven by iteration-to-iteration variance in text generation rather than by inter-class differences between high- and low-pollution days or between people with lung disease and healthy individuals.

\begin{figure*}[htpb!]
\centering
\includegraphics[width = 0.9\textwidth]{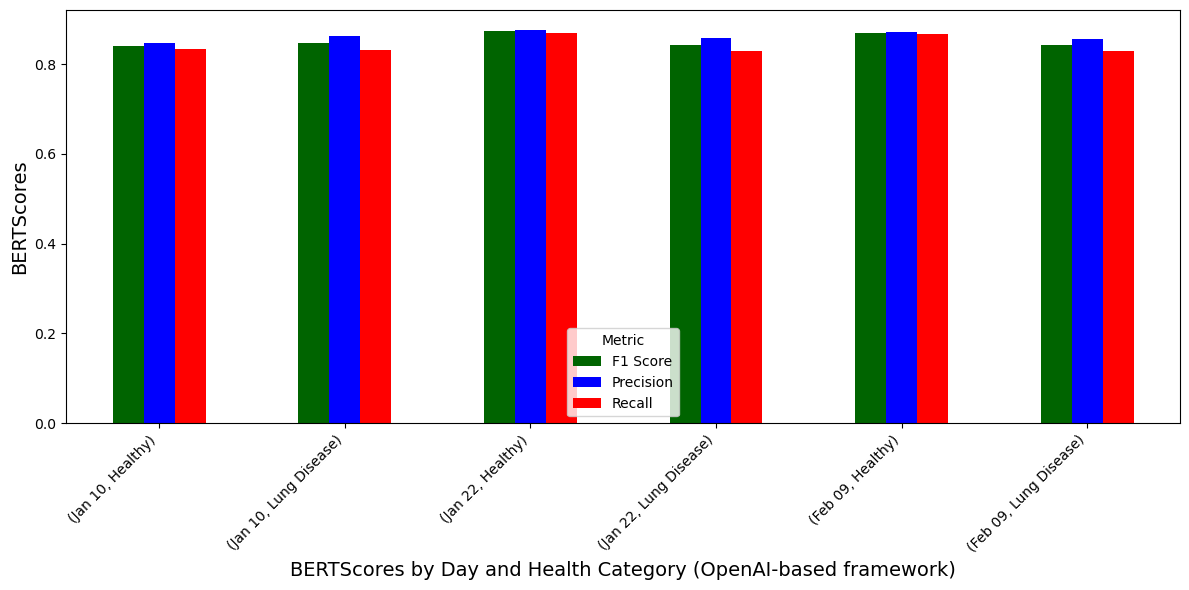}
\caption{Bar graph of BERTScores for health recommendations by day (air quality) and health class.}\label{fig:bert_barplots}
\end{figure*}

Moreover, the lack of outliers (which would have been plotted as circles) in Figure \ref{fig:box_plotsberta}indicates that the recommendations' semantic alignments were consistently maintained with the ground truths provided by Google's Air Quality API. The model is fairly consistent in generating good scores, as shown in Figure \ref{fig:bert_barplots}. This is also observed by examining the health recommendation with minimum BERTScores (precision = 0.808, recall = 0.778, F1 score = 0.793), which occurs at the fourth iteration for January 10 for people with lung disease, shown in Response \ref{verb:recommendations}. The scores show a positive semantic alignment, which is corroborated by comparing the response to the ground truth. 

By examining Response \ref{verb:recommendations}, we also note a substantial difference in writing style between the LLM system's responses (more concise) and the ground truth writing style (more elaborate). Aligning the writing style more closely with the ground truth would be trivially easy and would likely yield higher average BERT scores. However, this increase in score would not necessarily indicate better recommendations in practice. We believe the absence of outliers with inversely aligned semantics is a better indicator of recommendation success.

The reasoning model-based Instructor LLM is also capable of generating more complex and comprehensive reports and recommendations with cross-references to data. We performed preliminary tests of this scenario by prompting the system to generate a short report on the air quality. Without a ground truth policy to compare against, a numerical benchmark cannot be provided. However, visual inspection indicates that the example recommended policy report is reasonable. 
An example of a short policy recommendation report based on air quality data is shown below. The Instructor LLM was prompted with \textit{The area was affected by wildfires. Generate a short report with policy recommendations based on the air quality}.
\vspace{12pt}
\begin{lquote} Air Quality Report and Policy Recommendation
\hrule
1. \textbf{Severe Pollution Episodes}: Extreme PM2.5/PM10 spikes coincided with wildfires, far exceeding WHO guidelines.  
   - Example: Peaks of 452–509 µg/m³ on 01/10–01/12 (WHO 24-hour PM2.5 limit: 15 µg/m³).  

2. \textbf{Prolonged Exposure}: Elevated PM2.5/PM10 levels persisted for weeks, indicating chronic air quality degradation.  
   - Example: PM10 $>$70 µg/m³ on 01/21; PM2.5 $>$25 µg/m³ recurrently.  

3. \textbf{Recurring Outliers}: Hourly PM2.5/PM10 peaks suggest ongoing intermittent wildfire activity or secondary dust dispersion.  
   - Example: 30.1 µg/m³ PM2.5 on 01/31; 108.05 µg/m³ PM10 on 01/23.  

4. \textbf{Policy Recommendations}:  

   - \textbf{Emergency Response Framework}:  
     - Deploy real-time SMS/APP alerts prioritizing vulnerable groups.  
     - Distribute N95 masks and designate HEPA-filtered emergency shelters.  

   - \textbf{Wildfire Mitigation}:  
     - Implement pre-fire season controlled burns to reduce fuel loads.  
     - Invest in satellite/ground sensors for early wildfire detection.  

   - \textbf{Health Infrastructure}:  
     - Stockpile respiratory medications and expand ICU capacity.  
     - Launch public awareness campaigns on PM2.5 health risks.  

   - \textbf{Long-Term Air Quality Management}:  
     - Regulate agricultural/industrial burning during dry seasons.  
     - Establish green buffers with fire-resistant vegetation.  

   - \textbf{Climate Resilience}:  
     - Promote drought management policies to reduce wildfire risk.  
     - Strengthen cross-regional collaboration for air quality monitoring.  
     \hrule
\end{lquote}
\vspace{12pt}

We also tested against a ground truth official advisory warning from the Los Angeles Fire Department (LAFD) \citep{LAFD}. We used an OpenAI-based pipeline to generate 100 different health recommendations with an identical initial user prompt. We then generated 100 new health recommendations with a refined user prompt with additional context information (providing the Instructor LLM with information that 1) the pollution is generated by the Los Angeles Wildfires of January 2025 in the Palisades and Eaton Neighborhoods, and 2) the LLM is an AI assistant to the Los Angeles Fire Department and should assist in creating a recommendation message. The official LAFD warning is shown below.

\begin{quote}
\scriptsize
\begin{tcolorbox}[colback=white, colframe=black, boxrule=0.5pt, sharp corners]
\begin{lstlisting}[language=json]
"Residents are reminded that exposure to ash from the Palisades and Eaton wildfires remains a potential health concern and are reminded to take proper precautions to protect themselves. Windblown ash particles may be too large to be detected by air quality instrumentation and not reflected in the Air Quality Index (AQI) map. However, ash is typically visible to the naked eye either in the air or on outdoor surfaces. Although the AQI may show green, hazardous ash may be present in the air."
\end{lstlisting}
\end{tcolorbox}
\end{quote}

\begin{figure}[htpb]
\centering
        \includegraphics[width=0.45\textwidth]{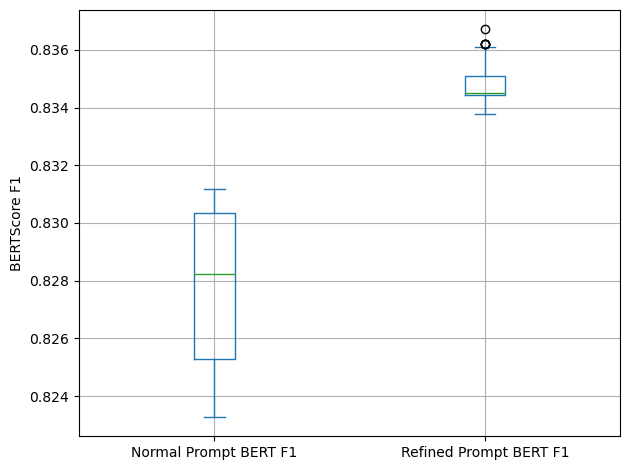} 
\caption{Boxplot of BERTScore F1 distribution comparing the ground truth Los Angeles Fire Department health recommendation to recommendations generated from our framework.}\label{fig:official}
\end{figure}

Figure \ref{fig:official} shows the BERTScore F1 comparison of both types of generated recommendations vs. LAFD recommendations. The results indicate 1) that both normal and refined recommendations show high sentiment overlap with the LAFD ground truth. 2) The refinement resulted in a statistically significant improvement, with the upper quartile of the normal recommendations lying below the lower quartile of the refined recommendations and the maximum of the normal recommendations lying below the minimum of the refined recommendations. 3) The effect of the refinement is small (despite being statistically significant) with the mean BERTScore F1 by 0.0068 after refinement. The minimum scores are 0.8233 for the normal prompt and 0.8337 for the refined prompt for comparison against the LAFD recommendations.

\subsection{Numerical and Statistical Summarization}\label{sec:4.3}
We prompted the LLM system to extract the daily mean and standard deviation of PM10 and PM2.5 levels from each data chunk. The workflow is similar to the previous health recommendation experiment; the user provides the initial prompt to the Instructor LLM, which provides the system prompts for each Worker LLM. We tested DeepSeek-V3, GPT-4o, which are fast autoregressive chat completion models, as well as GPT-o1 a slow reasoning model with stronger problem-solving skills. We omitted DeepSeek-R1 (DeepSeek's reasoning model) due to server-side service refusal.

\begin{figure*}[htbp]
    \centering
    \begin{subfigure}{0.45\textwidth}
        \centering
        \includegraphics[width=\textwidth]{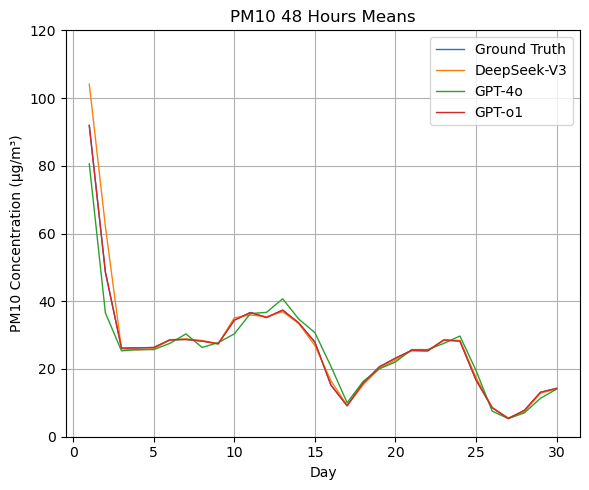}  
        \subcaption{Per-chunk 48 hours PM10 mean calculated from different Worker LLMs.}
    \end{subfigure}
    \begin{subfigure}{0.45\textwidth}
        \centering
        \includegraphics[width=\textwidth]{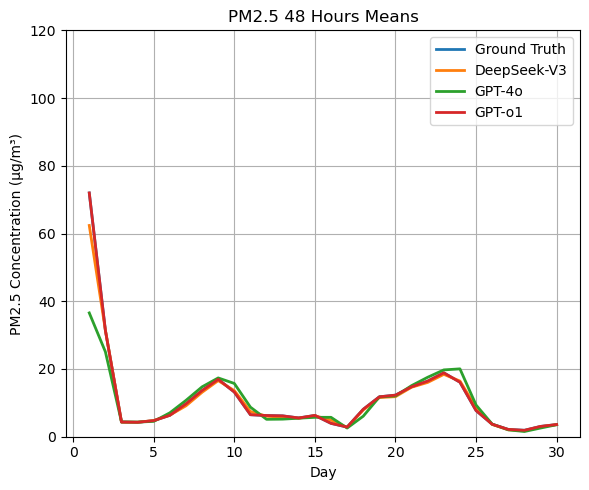}  
        \subcaption{Per-chunk 48 hours PM2.5 mean calculated from different Worker LLMs.}
    \end{subfigure}

        \begin{subfigure}{0.45\textwidth}
        \centering
        \includegraphics[width=\textwidth]{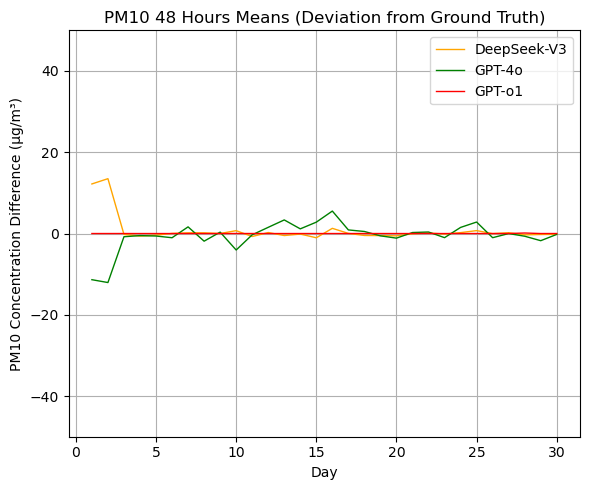}  
        \subcaption{PM10 means deviation from ground truth calculated from different Worker LLMs.}
    \end{subfigure}
    \begin{subfigure}{0.45\textwidth}
        \centering
        \includegraphics[width=\textwidth]{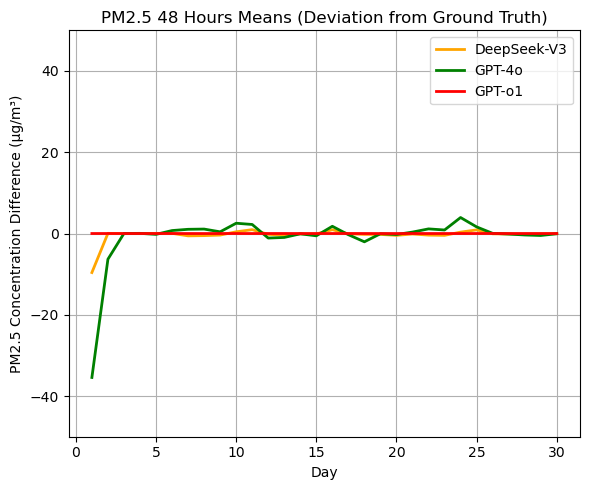}  
        \subcaption{PM2.5 means deviation from ground truth calculated from different Worker LLMs.}
    \end{subfigure}

        \caption{Per-chunk 48 hours means of PM10 and PM2.5 from Worker LLM numerical calculation, and the deviations of calculated means from ground truth means.} \label{Fig:stat_boxes}
\end{figure*}

{Average mean absolute error (MAE) and root-mean-squared error (RMSE) are shown in Table \ref{tab:worker_stats_metrics}. The results indicate that non-reasoning models struggle with arithmetic, with standard deviation calculations deviating from the ground truth more than mean calculations across all three models. GPT-o1, the reasoning model we tested, achieved a perfect 48-hour mean calculation but struggled with minor errors in standard deviation calculations. We also plotted the 48-hour means and the deviations of the calculated means from the ground truth in Figure \ref{Fig:stat_boxes}. The results show the deviations occur in the first chunk with the high-level spikes in PM10 and PM2.5 shown in Figure \ref{fig:pm_levels}.

\begin{table}[h]
\centering
\caption{MAE / RMSE values ($\mu g/m^3$) over 30 days for mean and standard deviations of PM levels across 48 hours as calculated by different worker LLMs.}
\begin{tabular}{lccc}
\toprule
Metric (MAE/RMSE) & DeepSeek-V3 & GPT-4o & GPT-o1 \\
\midrule
PM10 Mean     & 1.2 / 3.4   & 2.1 / 3.5   & 0.0 / 0.0 \\
PM10 Std      & 1.5 / 3.8   & 3.4 / 8.8   & 0.2 / 0.3 \\
PM2.5 Mean    & 0.6 / 1.8   & 2.2 / 6.7   & 0.0 / 0.0 \\
PM2.5 Std     & 0.9 / 2.6   & 4.7 / 14.1  & 0.1 / 0.3 \\
\bottomrule
\end{tabular}
\end{table}
\label{tab:worker_stats_metrics}

\subsection{Cost Estimation  and Scalability}
\red{The LLM model API call prices are displayed in Table \ref{Tab:modelcost}. Since the LLMs are hosted by their respective service providers, we do not incur hardware costs or runtime costs. Instead, there is a per-API call cost charged by the LLM providers and the data providers. These service providers allow for massive simultaneous user access. As such, for scaling our framework to a large user base, we estimate a typical per-use cost based on the API pricing instead of the hardware requirements. The following estimations are based on June 2025 pricing.}

\red{For our experiments, a typical 48-hour chunk of Google Maps Air Quality History response contains approximately 25{,}000 tokens, depending on the response format. Using a GPT-4.0/4.1 model to produce a per-chunk summary, which requires roughly 200 output tokens. This results in a cost of $
\frac{25{,}000}{1{,}000{,}000}\times2.00\$ + \frac{200}{1{,}000{,}000}\times8.00\$ = 0.05\$$
per chunk. The summarization of all 30 chunks results in a calculated cost of 1.50\$. The true cost is typically lower by a factor of 2 to 3 due to the discounted price of repetitive input (see input caching Table \ref{Tab:modelcost}). The response aggregation requires very few tokens in comparison. For example, analyzing a 200-token response from each of the 30 chunks to generate a 500-word report, or a 100-word policy recommendation, would require roughly 0.01\$ using a similar calculation, and is negligible compared to chunk summarization. As earlier suggested, it is viable to generate summaries using a cheaper model such as DeepSeek V3, which would reduce costs by roughly a factor of eight.}

\red{The retrieval of Google Maps Platform data can also incur a cost. Google Maps Platform offers a free 10{,}000 API calls for Air Quality Retrieval (and for various other data streams) with a cost of 5.00\$ per additional 1{,}000 calls. An hourly Air Quality History API call can retrieve up to 7 days of hourly air quality data in one API call. We did not exceed this free usage limit during our experiments and found the data retrieval cost to be negligible in standard use cases for each individual user.}

\red{If our framework or a similar implementation were to be deployed to a large user base, both LLM API calls and data retrieval API calls costs may accumulate rapidly. In this situation, we recommend offloading each user's cost to the user or considering commercialization options with more detailed per-use calculations. If deploying the framework locally, a large user base would also incur infrastructure costs for receiving user requests and processing user requests, and incur hardware costs for tasks with previously negligible hardware requirements, such as prompt and and message storing/passing when running the framework. Unless the framework's host already has powerful computing hardware and network infrastructure (e.g., a university or a corporation), we recommend deploying the framework using cloud-computing platforms that have linear or sublinear cost scaling with the number of users.}

\begin{table}[ht]
\centering
\captionof{table}{API call cost* of relevant OpenAI and DeepSeek models}
\begin{tabular}{l|c|c}
\hline
Model Class & \shortstack{Input Price**\\/1M Token (USD)} & \shortstack{Output Price\\/1M Tokens (USD)} \\ \hline
GPT4.1${^1}$                      & 2.00 (0.50)             & 8.00              \\
GPT4.1${^1}$ Mini                & 0.40 (0.10)             & 1.60               \\
Deepseek V3               & 0.27 (0.07)             & 1.10               \\
Deepseek R1     & 0.55 (0.14)              & 2.19               \\
OpenAI-o3${^2}$      & 2.00 (0.50)            & 8.00              \\ \hline
\end{tabular}
\par
\smallskip
\justifying
\noindent
*Table compiled on June 27, 2025. Prices and model deployments are subject to changes by their respective service providers. Although we used ChatGPT4o and GPT-o1 in our experiments, these were replaced with ${^1}$GPT4.1 and ${^2}$OpenAI-o3 models by the service provider at the time of compiling this table.\\
**Both DeepSeek and OpenAI provide an automatic discount for multiple calls with repetitive input (i.e., input caching). These discounts are displayed inside parentheses.
\label{Tab:modelcost}
\end{table}

\subsection{Discussion}
\red{In terms of general limitations, LLMs are prone to hallucinations and lack the appropriate context for specific use cases (e.g., for disaster response). LLMs without the use of external tools also currently struggle with complex numerical calculations. Although external tools can be automatically called to assist numerical calculations, and additional context can be provided (as shown in Figure \ref{fig:official}) to improve policy alignment, results are nonetheless imperfect. As such, LLMs should only be used as assistants to human experts for policy and report generation, if at all.}

\red{In terms of ethical concerns, the creation of a technical/scientific report or a policy recommendation is a high-impact task. LLM-generated content, in this case, raises concerns for reliability, accountability, data privacy, and manipulation risks. As mentioned, the accuracy and alignment of generated content are imperfect. For high-impact tasks, this naturally raises ethical concerns when errors in the generated content cause real damage. This directly translates into an accountability concern if errors from LLM-generated technical, scientific, or policy content occur, as there is no clear actor to bear responsibility for damages caused by such errors. For certain use cases, AI-generated content can also be considered fraudulent. Additionally, in some modes of use, such as the ones presented in this paper, data is sent to the servers of LLM providers through API calls. This raises ethical concerns for sensitive/private data. Finally, the training and deployment of LLMs are opaque and prone to manipulation by the provider. Pretrained LLMs may be conditioned to respond in specific manners on specific topics without the user's knowledge, resulting in biases. This may be unintentional, due to existing bias in large-scale internet-crawled datasets used to train LLMs, or intentional due to the LLM providers' designs. Moreover, the LLM service provider can add hidden prompts and rules that guide LLM responses on specific topics. As such, LLM users in general should be aware of the limitations and ethical concerns of using LLMs for high-impact use cases, including report generation and policy recommendations.}

\red{Google Maps Platform offers extensive data coverage. With few exceptions, data from most regions in most countries can be retrieved by simply using the geographic coordinates or other geocoding information. As such, our framework can easily analyze the air quality data in other use cases beyond this LA case study. Moreover, Google Maps Platform offers other data streams, including weather, solar potential, traffic, and many more. Aside from our system prompts, the LLMs and the framework itself are also not specific to wildfire and air quality analysis; as such, our framework can easily be extended to analyze other GIS data. And finally, the integration with Google Maps Platform itself is modular, and another online GIS platform's data streams can easily be integrated instead.}

Although our framework was built and tested for OpenAI and Deepseek APIs, we only benchmarked OpenAI models for repeated BERTscore assessments. Although Deepseek is a less expensive LLM service provider, as of February 2025, the platform is less reliable. When attempting to make continuous API calls to Deepseek, we quickly ran into service refusals/failures to respond. To allow for more LLM models and provider choices, we plan to integrate LLM APIs from other LLM service providers such as Google's Gemini API, Anthropic's Claude API, and xAI's Grok API.

We also would like to address the lack of spatial analysis. This is due to two factors: 1) adding a spatial dimension increases the cost unless we design a new system for spatial analysis (which is outside the scope of our current LLM system). By simply increasing the number of coordinates at which we retrieve the data and increasing the number of LLM API calls accordingly, we quickly run into a cost overrun. Secondly, Google Maps Platform only stores air quality data for 30 days. When we wished to test the spatial distribution of the data after the initial experiments, the data had already become inaccessible. Adding spatial analysis capabilities is a future research direction we wish to pursue.

The high variance in responses generated by LLMs and LLM systems indicates that benchmarking LLMs and LLM systems should be done with great care. When single samples are used for score assessment, it is possible that the variance in the response out-scales any contributions from proposed innovations.  We believe a new metric should be developed for policy recommendation since full-reference text assessment metrics are too influenced by writing style alignment, whereas a measure of the semantic anti-alignment is more important for practical purposes. 

Additionally, we believe a more rigorous way to perform statistical analysis using LLMs should be developed. Although the analysis generated by the LLM system in Section \ref{multi_agent_analysis} can suffice for some purposes, it does not have scientific rigor. Our preliminary experiments in February 2025 showed that the tested OpenAI GPT-o1 and Deepseek-R1 models still struggled with hypothesis testing and with calculating non-elementary mathematical operations and functions, as well as predictive modeling. The experiments in Section \ref{sec:4.3} further show that simpler LLMs are especially prone to arithmetic errors. They should not be used naively for predictive mathematical modeling (i.e. time series prediction) and predictive analysis outside of simple trends. We believe an LLM system that can perform rigorous hypothesis testing and provide $p$-value estimates should be designed, perhaps by leveraging LLM systems that call external statistical libraries. This will greatly improve data analysis efficiency in experiments in many fields of science.


\section{Conclusion}
Building on our previous work, we created an LLM System with Instructor-Worker Architecture capable of cloud platform data retrieval and data analysis. We tested the capability of our system by analyzing Google Maps Platform's air quality data for the Los Angeles wildfires of January 2025. This research showed the viability of our system for data analysis and policy recommendations. The system was able to identify spike events in trends and outliers in air quality, provide health recommendations, and provide summary reports. The recommended policies were always aligned with ground truth policies with a minimum BERTScore precision of 0.808, and BERTScore recall of 0.778. Comparisons to the official LAFD health warning show an even better semantic alignment with BERTScore F1 always above 0.820, especially when the framework is provided with a more detailed context. The generated recommendations were always of the correct sentiment and semantics. However, experiments on using LLMs to extract statistical information in a verbose dataset show that even powerful reasoning models still struggle with numerical computations. Unlike supervised models, our framework can be directly applied to new out-of-distribution data without ground truth, and can help perform automated analysis of cloud-connected air quality data streams, assisting wildfire study, research, and monitoring.



%

\printcredits

\section*{Declaration of competing interest}
The authors declare that they have no known competing financial interests or personal relationships that could have appeared to influence the work reported in this paper.

\section*{Acknowledgements}
This work was supported in part by the Natural Sciences and Engineering Research Council of Canada (NSERC) Discovery Grant (No. RGPIN-2022-03741).

\section*{\red{Data Availability}}
\red{Data will be made available on request.}

\bibliographystyle{cas-model2-names}
\bibliography{References}

\newpage

 




\vfill

\end{document}